\documentclass[10pt,twocolumn,letterpaper]{article}

\usepackage{cvpr}              % To produce the CAMERA-READY version
\usepackage[accsupp]{axessibility} 

\usepackage{graphicx}   % For \includegraphics and \rotatebox
\usepackage{multirow}   % For \multirow
\usepackage{array}      % For column formatting like >{\centering...}
\usepackage{booktabs} % For professional lines like \toprule
\usepackage{siunitx}  % For S column type (aligning numbers by decimal point)
\usepackage{graphicx}

\definecolor{cvprblue}{rgb}{0.21,0.49,0.74}
\usepackage[pagebackref,breaklinks,colorlinks,allcolors=cvprblue]{hyperref}

\def\paperID{} % *** Enter the Paper ID here
\def\confName{CVPR}
\def\confYear{2026}

\title{HiFICL: High-Fidelity In-Context Learning for Multimodal Tasks}

\author{
    \makebox[\linewidth][c]{
        \setlength{\tabcolsep}{3pt}
        \begin{tabular}{cccc}
            Xiaoyu Li$^{1,2}$\thanks{Equal contribution.} & Yuhang Liu$^1$\footnotemark[1] & Xuanshuo Kang$^1$ & Zheng Luo$^1$ \\
            {\tt\scriptsize xiaoyuuestc@uestc.edu.cn} & {\tt\scriptsize 202422090537@std.uestc.edu.cn} & {\tt\scriptsize 202422090729@std.uestc.edu.cn} & {\tt\scriptsize 202421090326@std.uestc.edu.cn}
        \end{tabular}
    }
    \\[0.3cm]
    \makebox[\linewidth][c]{
        \setlength{\tabcolsep}{8pt}
        \begin{tabular}{ccc}
            Fangqi Lou$^1$ & Xiaohua Wu$^{1,2}$\thanks{Corresponding author.} & Zihan Xiong$^1$ \\
            {\tt\scriptsize 202422090730@std.uestc.edu.cn} & {\tt\scriptsize wuxh@uestc.edu.cn} & {\tt\scriptsize 2022090905007@std.uestc.edu.cn}
        \end{tabular}
    }
    \\[0.4cm]
    $^1$University of Electronic Science and Technology of China \\
    $^2$Institute of Electronics and Information Industry Technology of Kashgar
}

\begin{document}
\maketitle 

\begin{abstract}
    In-Context Learning (ICL) is a significant paradigm for Large Multimodal Models (LMMs), using a few in-context demonstrations (ICDs) for new task adaptation. 
    However, its performance is sensitive to demonstration configurations and computationally expensive. 
    Mathematically, the influence of these demonstrations can be decomposed into a dynamic mixture of the standard attention output and the context values. 
    Current approximation methods simplify this process by learning a ``shift vector''. 
    Inspired by the exact decomposition, we introduce \textbf{Hi}gh-\textbf{F}idelity \textbf{I}n-\textbf{C}ontext \textbf{L}earning (\textbf{HiFICL}) to more faithfully model the ICL mechanism. 
    HiFICL consists of three key components: 1) a set of ``virtual key-value pairs'' to act as a learnable context, 2) a low-rank factorization for stable and regularized training, and 3) a simple end-to-end training objective. 
    From another perspective, this mechanism constitutes a form of context-aware Parameter-Efficient Fine-Tuning (PEFT). 
    Extensive experiments show that HiFICL consistently outperforms existing approximation methods on several multimodal benchmarks. The code is available at \url{https://github.com/bbbandari/HiFICL}.
 \end{abstract}    
\section{Introduction}
\label{sec:intro}

% 【排版核心技巧】：把跨栏大图直接提到整个章节的最前面！
% 这样 LaTeX 在处理第一页的第一段时就能把图排进队列，确保它出现在第二页顶部。
\begin{figure*}[t]
  \centering
  \includegraphics[width=\linewidth]{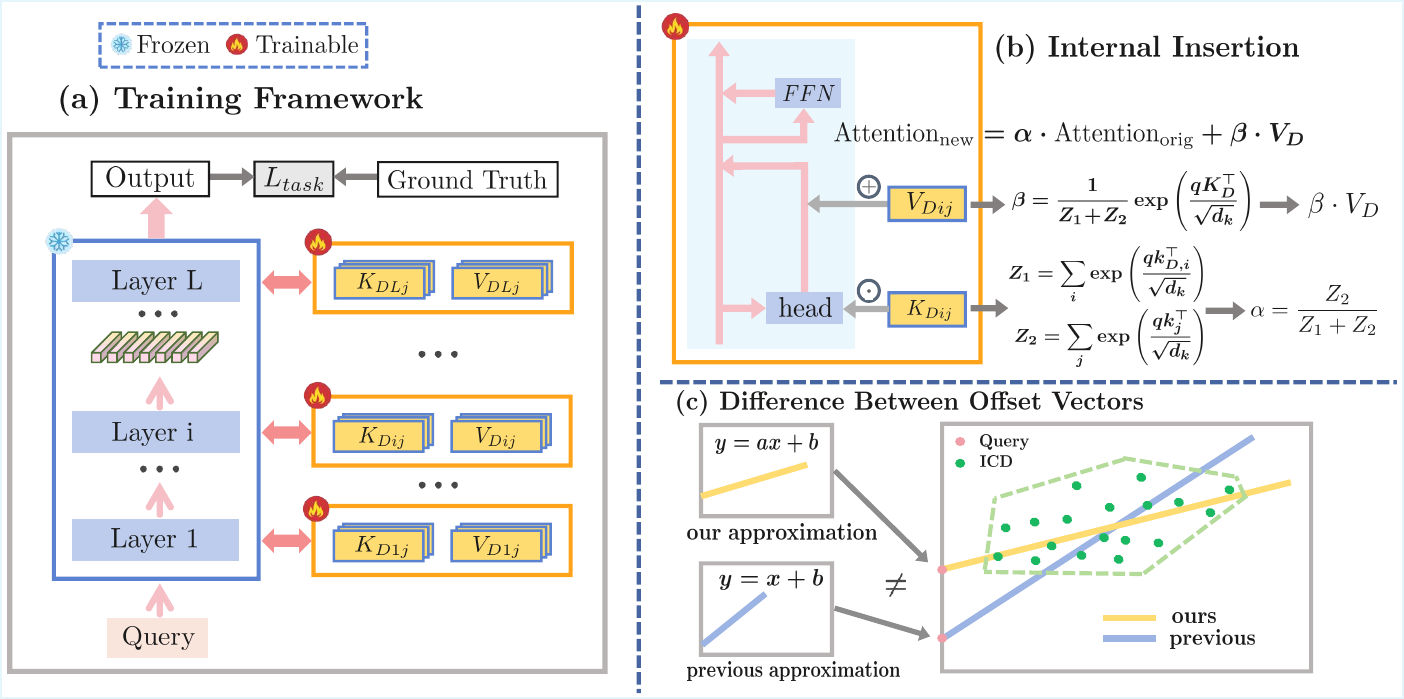}
  \caption{Overview of the HiFICL framework. (a) The overall training process involves a frozen LMM backbone and layer-wise trainable virtual key-value pairs, optimized via a final task loss. (b) Inside each attention head, these virtual pairs are used to compute the high-fidelity attention output according to our derived formula (\cref{eq:decomposition}). (c) Conceptually, unlike previous linear shift methods that only learn an additive offset ($y = x + b$), HiFICL introduces a dynamic scaling factor ($\alpha$) to modulate the original attention ($y = ax + b$). This non-linear partitioning provides a more faithful fit to the underlying data manifold.}
  \label{fig:overview}
\end{figure*}

In-Context Learning (ICL) has become a key capability of Large Language Models (LLMs) and Large Multimodal Models (LMMs), enabling them to perform novel tasks by conditioning on a few examples without parameter fine-tuning~\cite{brown2020language, Alayrac2022Flamingo}. This powerful few-shot paradigm is finding rapidly expanding applications across complex domains, including nuanced visual reasoning~\cite{hurst2024gpt, zhang2021mme}, robotic control from visual prompts~\cite{zitkovich2023rt, driess2023palm}, and even medical image analysis~\cite{tu2024towards, li2023llava}, where adapting to novel instruments or patient data is crucial. However, the practical utility of ICL, particularly in the multimodal space, is severely constrained by two fundamental challenges. First, the high token cost of visual inputs leads to prohibitive computational overhead and restricts the number of in--context demonstrations (ICDs)~\cite{liu2023visual}. Second, ICL performance exhibits high sensitivity to the selection and ordering of these demonstrations, undermining its reliability~\cite{li2024configure, lu2021fantastically}.

To mitigate these issues, a dominant research direction has focused on \textit{approximating} the ICL effect. This paradigm, centered on learning a ``shift vector,'' aims to distill the knowledge from explicit ICDs into a compact representation that can be efficiently injected into the model~\cite{hendel2023context, todd2023function, peng2024live, jiang2025mimic, li2025m2iv}. While this approach has proven effective in reducing inference costs, we argue that this paradigm is predicated on a simplified and potentially restrictive linear assumption. These methods treat the ``shift'' as an external, additive component to be learned, focusing on approximating the emergent outcome of the ICL process. This overlooks a more fundamental question: what is the underlying causal mechanism that generates this effect in the first place?

In this paper, we challenge this foundational premise. Our analysis (\Cref{sec:methodology}) reveals the ``shift effect'' is not an approximation target but a direct analytical consequence of the attention formula, precisely equivalent to the dynamically-weighted sum of the context value matrix ($V_D$). Prior methods were effectively approximating an outcome whose exact form was already embedded in the original equation. This insight reframes the problem entirely, shifting the objective from approximating the \textit{effect} to efficiently parameterizing the \textit{source} ($K_D, V_D$). This clarification, we argue, advances the long-standing discussion on the ``shift vector'' to a more principled foundation.

Based on this new understanding, we introduce \textbf{Hi}gh-\textbf{F}idelity \textbf{I}n-\textbf{C}ontext \textbf{L}earning (\textbf{HiFICL}). HiFICL abandons the indirect simulation of the shift vector and instead directly parameterizes the source of ICL within the attention module itself. This is achieved through a set of learnable, low-rank ``virtual key-value pairs,'' which act as a compact, task-adaptive memory. These virtual demonstrations interact with the query dynamically through the native softmax computation, faithfully emulating the role of explicit ICDs and preserving the inherent non-linearity of the attention mechanism. An overview of our framework is presented in \cref{fig:overview}.

This reframing of the ICL approximation problem aligns with a growing body of research questioning the adequacy of the linear shift model. Recent studies have begun to theorize that ICL itself functions as a form of ``inference-time finetuning,'' where the model's behavior is dynamically and temporarily optimized by the provided context, much like a few steps of gradient descent~\cite{dai2022can, li2024nonlinear}. This perspective, supported by empirical evidence showing a striking similarity between the effects of ICL and finetuning on model confidence~\cite{anonymous2025detecting}, suggests that the ICL mechanism is fundamentally a learning process, not a simple vector addition. While these works provide the theoretical and empirical grounding for this hypothesis, our work provides the first concrete instantiation of it as a \textit{training-time} PEFT method. HiFICL is designed to distill and solidify this transient, inference-time adaptation into a permanent and efficient set of trainable parameters. Viewed through this lens, its mechanism presents a more logically coherent form of adaptation than conventional methods like LoRA~\cite{hu2022lora}, which apply static, input-agnostic updates. By directly simulating the foundational process of ICL through learnable, context-aware modulation, HiFICL offers a dynamic and principled approach to model adaptation.

Our contributions are summarized as follows:
\begin{itemize}
    \item We reframe the ICL approximation problem by providing a rigorous mathematical derivation that shifts the objective from approximating an indirect \textit{effect} to directly parameterizing its \textit{source}. This analysis also theoretically clarifies the underlying mechanism of ICL within the attention formula.
    \item We propose HiFICL, a novel and practical framework that materializes this principle. HiFICL employs a dual low-rank virtualization and a teacher-free, end-to-end objective to achieve a new form of dynamic, context-aware PEFT.
    \item We conduct extensive experiments showing that HiFICL consistently establishes a new state-of-the-art on multiple authoritative multimodal benchmarks, empirically validating the superiority of our high-fidelity paradigm.
\end{itemize}
\section{Related work}
\label{sec:related_work}

\subsection{ICL in Large Multimodal Models}

In-Context Learning (ICL) enables few-shot adaptation without parameter updates in Large Language Models (LLMs)~\cite{brown2020language, dong2022survey}. This paradigm has naturally extended to Large Vision-Language Models (LVLMs), where early works established the viability of interleaved image-text conditioning~\cite{Alayrac2022Flamingo, awadalla2023openflamingo, laurenccon2023obelics, tsimpoukelli2021multimodal}. However, multimodal ICL introduces unique challenges compared to the text domain~\cite{luo2025robobench}, as models often neglect visual contexts~\cite{chen2025can}. First, the high token cost of visual inputs drastically limits context windows and incurs prohibitive computational overhead~\cite{liu2023visual, xiao2023efficient}. Second, cross-modal synergies amplify performance sensitivity to demonstration selection and ordering~\cite{li2024configure, lu2021fantastically}. Unlike in LLMs, visually similar retrievals may misalign with task semantics in LMMs~\cite{yang2024lever}, complicating optimal configuration searches in complex feature spaces~\cite{qin2024factors, yang2023exploring}. While recent works attempt sequence optimization~\cite{li2025taco} or attention modulation~\cite{li2025cama, li2025towards, chen2025true} to mitigate instability, the fundamental hurdles of computational cost and extreme sensitivity remain, strongly motivating the paradigm of ICL approximation.

\subsection{The Linear Shift Approximation}
To address the limitations of vanilla ICL, a dominant paradigm has emerged that approximates its effect via ``representation engineering.'' Early non-learnable methods proposed abstracting demonstrations into a single vector, such as a ``Task Vector''~\cite{hendel2023context} or a ``Function Vector''~\cite{todd2023function}, and adding it to the model's hidden states. While insightful, these static methods proved insufficient for the complexity of cross-modal tasks. This spurred the development of learning-based approaches that learn a ``shift vector.'' Key works like LIVE~\cite{peng2024live} and MimIC~\cite{jiang2025mimic} introduced sophisticated, query-dependent vectors learned via knowledge distillation and inserted them directly into Transformer layers. Despite architectural refinements, these methods are unified by a shared, simplifying assumption: they model the complex ICL effect as a post-hoc, additive linear shift to hidden state representations.

However, recent advances in mechanistic interpretability reveal a fundamental contradiction with this linear paradigm. Research has shown that ICL is not a simple global shift but is executed by specialized circuits. A key discovery is the role of ``Induction Heads''~\cite{olsson2022context, elhage2021mathematical}, attention heads that perform complex pattern matching and retrieval. These mechanisms are believed to be how models implicitly implement learning algorithms like gradient descent~\cite{dai2022can, wang2023label}. Furthermore, analyses from the perspective of hidden state geometry demonstrate that ICL is a highly non-linear transformation that dynamically reshapes the representation space to enhance class separability and align with decision boundaries~\cite{yang2025unifying}. This difference between the linear assumption of current approximation methods and the non-linear reality of the underlying mechanism suggests that relying solely on the linear shift paradigm may restrict the model's full potential, motivating the exploration of more faithful approximation models.

\subsection{Low-Rank Adaptation for PEFT}

The connection between In-Context Learning (ICL) and finetuning is not merely a conceptual analogy but is rooted in the mechanistic underpinnings of the Transformer architecture. A growing body of theoretical research posits that the self-attention mechanism can implicitly simulate optimization algorithms. These studies show that attention heads can identify relevant context examples and aggregate their label information in a manner that mathematically mirrors one or more steps of gradient descent on a prediction loss~\cite{dai2022can, li2024nonlinear, ren2024towards}. This frames ICL as a form of ``inference-time'' optimization, where the model effectively fine-tunes its internal representations dynamically for each query.

This raises a natural question: how does this implicit, dynamic optimization relate to explicit finetuning? Parameter-Efficient Fine-Tuning (PEFT)~\cite{he2021towards}, notably Low-Rank Adaptation (LoRA)~\cite{hu2022lora}, represents the dominant paradigm for the latter, injecting trainable low-rank matrices into frozen layers~\cite{dettmers2023qlora}. However, a core characteristic of LoRA is that its adaptation is static and input-agnostic. This creates a conceptual gap: while ICL performs a dynamic, per-query optimization, LoRA performs a static, global one. Our work is motivated by bridging this gap, aiming to create a PEFT method that is not only parameter-efficient but also retains the dynamic, context-aware adaptation properties inherent to the ICL mechanism.
\section{Methodology}
\label{sec:methodology}

Our methodology is built upon a foundational reframing of the In-Context Learning (ICL) approximation problem. We shift the objective from modeling the indirect \textit{effects} of ICL to directly parameterizing its \textit{source} within the attention mechanism. We begin by deconstructing the attention formula to derive the exact mathematical form of ICL's influence, establishing a rigorous theoretical foundation for our approach (\cref{sec:deconstruction}). Based on this analysis, we introduce HiFICL, a novel framework designed for high-fidelity ICL approximation (\cref{sec:hificl_design}). We then conclude by situating HiFICL within the broader landscape of model adaptation, clarifying its unique position by contrasting it with prior ICL approximators and static PEFT methods like LoRA (\cref{sec:hificl_vs_mimic} and \cref{sec:hificl_as_peft}).

\subsection{Mathematical Analysis}
\label{sec:deconstruction}

To construct a high-fidelity approximation model, we first return to the foundational principles of the Transformer attention mechanism. We derive the exact mathematical form of the attention output when In-Context Demonstrations (ICDs) are present. Consider an input sequence formed by concatenating demonstration examples $X_D$ and a query $X_q$. After projection by the weight matrices ($W_k, W_v$), this yields the corresponding key and value matrices: $K_{context} = [K_D, K_q]$ and $V_{context} = [V_D, V_q]$. For simplicity, we hereafter denote the query-derived keys and values, $K_q$ and $V_q$, as $K$ and $V$, respectively.

For any given query vector $q$ (a row from the query matrix $Q$), the output computation within a single attention head can be progressively decomposed as follows:
\begin{equation}
\label{eq:decomposition}
\begin{aligned}
& \text{Attn}_{\text{out}}(q, [K_D, K], [V_D, V]) \\
&= \text{softmax}\left( \frac{q[K_D, K]^\top}{\sqrt{d_k}} \right) \begin{bmatrix} V_D \\ V \end{bmatrix} \\
&= \left[ \frac{\exp\left( \frac{qK_D^\top}{\sqrt{d_k}} \right)}{Z_1 + Z_2}, \frac{\exp\left( \frac{qK^\top}{\sqrt{d_k}} \right)}{Z_1 + Z_2} \right] \begin{bmatrix} V_D \\ V \end{bmatrix} \\
&= \frac{Z_2}{Z_1 + Z_2} \cdot \frac{\exp\left( \frac{qK^\top}{\sqrt{d_k}} \right)}{Z_2} V + \frac{1}{Z_1 + Z_2}\exp\left( \frac{qK_D^\top}{\sqrt{d_k}} \right) V_D \\
&= \frac{Z_2}{Z_1 + Z_2} \text{softmax}\left(\frac{qK^\top}{\sqrt{d_k}}\right) V \\
&\quad + \frac{1}{Z_1 + Z_2} \sum_{i} \exp\left(\frac{qk_{Di}^\top}{\sqrt{d_k}}\right) v_{Di} \\
&= \alpha \cdot \text{SA}(q, K, V) + \beta \cdot V_D
\end{aligned}
\end{equation}
where $\text{SA}(q, K, V)$ is the standard self-attention output over the query tokens and $V_D$ can be viewed as the redefined offset vector basis. The coefficients $\alpha(q, K, K_D)$ and $\beta(q, K, K_D)$ are dynamically computed scalar and vector weights, respectively, defined as:
\begin{equation}
\label{eq:coefficients}
\alpha = \frac{Z_2}{Z_1 + Z_2}, \quad \beta = \frac{1}{Z_1 + Z_2}\exp\left( \frac{qK_D^\top}{\sqrt{d_k}} \right)
\end{equation}
Here, $Z_1 = \sum_{i} \exp(\frac{qk_{Di}^\top}{\sqrt{d_k}})$ and $Z_2 = \sum_{j} \exp(\frac{qk_j^\top}{\sqrt{d_k}})$ represent the sums of exponentiated, unnormalized attention scores between the query $q$ and the demonstration keys ($K_D$) and query keys ($K$), respectively.

\Cref{eq:decomposition} provides a crucial theoretical insight. It reveals that the effect of ICL is not a simple, externally added vector. Instead, it is an exact, analytical consequence of the attention formula, manifesting as a dynamic mixture of two components: 1) the standard self-attention output, scaled by the query-dependent coefficient $\alpha(q)$, and 2) the demonstration value matrix $V_D$, dynamically weighted by the vector $\beta(q)$. This formulation not only makes it clear that to faithfully model ICL, one must capture this entire dynamic system, but also theoretically clarifies the underlying mechanism of ICL within the attention formula itself. This stands in contrast to prior paradigms that oversimplify this process. Importantly, our derivation is grounded in unified self-attention architectures~\cite{liu2023visual}, not earlier cross-attention designs~\cite{Alayrac2022Flamingo}, aligning our method with the current state-of-the-art paradigm in LMMs.

\subsection{HiFICL: Direct Parameterization}
\label{sec:hificl_design}

Based on the theoretical insights from \cref{sec:deconstruction}, we introduce HiFICL. The core idea is to directly parameterize the source of ICL, the unknown demonstration key-value pairs $(K_D, V_D)$, rather than approximating its ultimate effect. We achieve this by introducing a set of virtual context slots, which manifest directly as ``virtual key-value pairs'' $(K_{learn}, V_{learn})$ within the attention mechanism. To capture the functional specialization of different attention heads~\cite{clark2019does, voita2019analyzing}, our modifications are applied on a per-head basis. We equip each attention head $h$ with an independent set of virtual pairs, $(K_{learn}^{(h)}, V_{learn}^{(h)})$, allowing each head to learn the specialized contextual information it requires for its sub-task.

Directly learning full-rank virtual matrices would introduce an excessive number of parameters, posing significant risks of overfitting and training instability. To address this, we leverage principles from Parameter-Efficient Fine-Tuning (PEFT) and propose a Dual Low-Rank Decomposition strategy, inspired by LoRA~\cite{hu2022lora}:
\begin{equation}
\label{eq:lora}
K_{learn}^{(h)} = K_A^{(h)} K_B^{(h)}, \quad V_{learn}^{(h)} = V_A^{(h)} V_B^{(h)}
\end{equation}
where $n$ denotes the sequence length of these virtual matrices, $K_A^{(h)}, V_A^{(h)} \in \mathbb{R}^{n \times r}$ and $K_B^{(h)}, V_B^{(h)} \in \mathbb{R}^{r \times d_h}$, with the rank $r \ll d_h$. This design elegantly provides two critical benefits. First, for structural stability, we initialize the $V_B^{(h)}$ matrices to zero. This guarantees that the contextual shift term in Eq.~(\ref{eq:decomposition}) is null at the beginning of training, preventing potential gradient explosion and creating a smooth learning trajectory from the base model's state. Second, the low-rank factorization of $K_{learn}^{(h)}$ acts as a powerful form of structural regularization. It compels the model to learn a compact and generalizable set of ``prototype keys,'' effectively creating an information bottleneck that mitigates overfitting.

To maximize the potential of this high-fidelity architecture, we adopt a simple yet powerful teacher-free, end-to-end optimization strategy. We depart from the teacher-student paradigm used in prior work~\cite{jiang2025mimic}, which relies on intermediate alignment losses. Instead, we optimize all trainable parameters, including the injected low-rank virtual matrices, using only the final supervised task loss. For generative tasks, this is typically the cross-entropy loss over the target sequence:
\begin{equation}
\label{eq:loss}
\mathcal{L}_{\text{task}} = -\sum_{t=1}^{T} \log P(A_{t} | Q, A_{<t} ; \Theta_{\text{base}}, \Theta_{\text{HiFICL}})
\end{equation}
where $\Theta_{\text{base}}$ are the frozen base model parameters and $\Theta_{\text{HiFICL}}$ represents all our trainable virtual parameters. This end-to-end approach grants the model full autonomy to learn optimal configurations for its virtual key-value pairs, guided solely by the ultimate task objective, thereby unlocking its full performance potential.

\subsection{Architectural Comparison}
\label{sec:hificl_vs_mimic}

\begin{figure}[t]
  \centering
  \includegraphics[width=\linewidth]{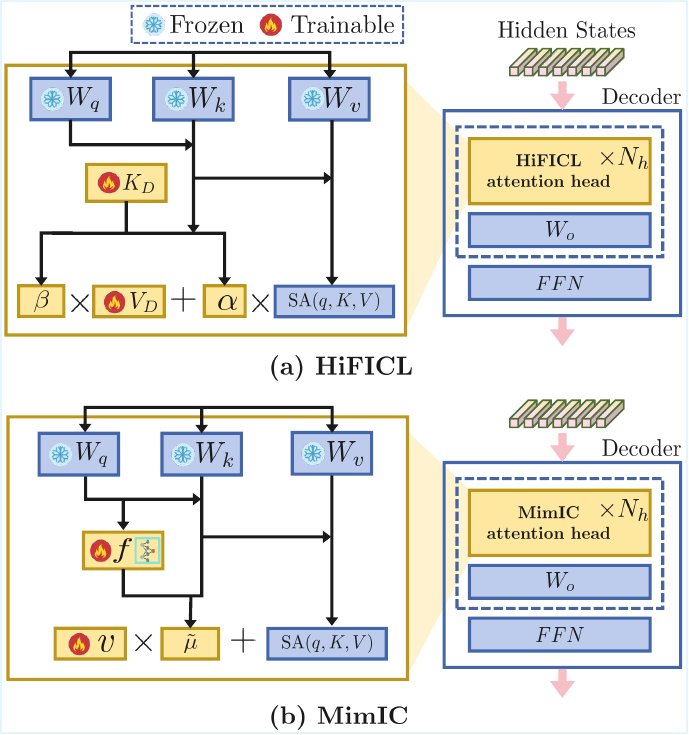}
  \caption{Architectural comparison of HiFICL and MimIC. (a) \textbf{HiFICL} implements the full, non-linear attention dynamic. (b) \textbf{MimIC} simplifies the ICL effect into a uni-directional linear shift.}
  \label{fig:hificl_vs_mimic}
\end{figure}

We contrast HiFICL with the prior state-of-the-art, MimIC~\cite{jiang2025mimic}, highlighting two fundamental advantages. \textbf{First}, in terms of model architecture, MimIC simplifies the ICL effect into a linear shift, learning a head-specific fixed direction and a query-dependent dynamic magnitude (\cref{fig:hificl_vs_mimic}b). In contrast, HiFICL is designed for higher fidelity; by parameterizing a contextual basis space with learnable virtual key-value pairs, it faithfully implements the complete, multi-directional, and non-linear dynamic mixture derived in our analysis (\cref{fig:hificl_vs_mimic}a). \textbf{Second}, in terms of training paradigm, HiFICL utilizes a simple and efficient end-to-end objective. This differs from MimIC, which relies on a costly teacher-student paradigm that requires aligning hidden states at each layer. As our ablations will show (\cref{sec:ablations}), this teacher-based approach not only incurs significant computational overhead but also acts as a performance ceiling, limiting the model's full potential.

% --- FIGURE for HiFICL vs. MimIC Comparison ---

\subsection{HiFICL as a Dynamic PEFT}
\label{sec:hificl_as_peft}

\begin{figure}[t]
  \centering
  \includegraphics[width=0.9\linewidth]{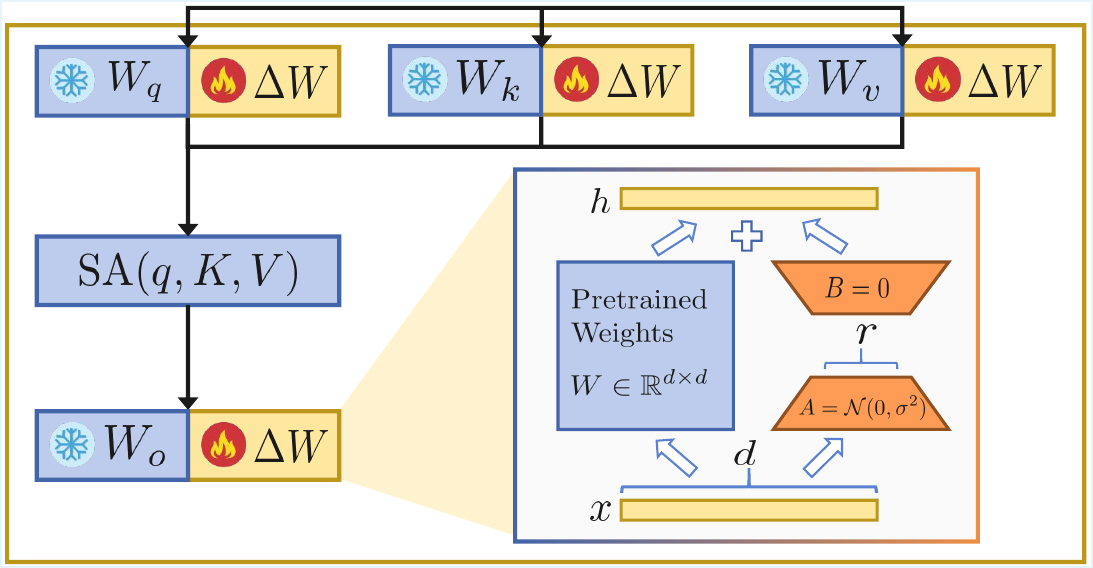}
  \caption{Illustration of the Low-Rank Adaptation (LoRA) mechanism. LoRA injects static, trainable low-rank updates ($\Delta W$) into the frozen weight matrices of an attention module.}
  \label{fig:lora_mechanism}
\end{figure}

Beyond ICL approximation, HiFICL introduces a new perspective on Parameter-Efficient Fine-Tuning (PEFT). As theorized, ICL functions as ``inference-time finetuning,'' dynamically optimizing model behavior via context~\cite{li2024nonlinear, anonymous2025detecting}. Our work provides the first concrete instantiation of this hypothesis as a training-time PEFT method. Contrasting HiFICL with Low-Rank Adaptation (LoRA)~\cite{hu2022lora}, a dominant technique that performs a static, input-agnostic adaptation in the weight space (\cref{fig:lora_mechanism}), HiFICL operates in the activation space, performing a dynamic, content-aware adaptation by providing the model with a learnable memory of virtual contextual examples. Its mechanism is thus more interpretable, analogous to teaching a model how to reason with context rather than applying static, input-agnostic modifications to its weights. As our experiments will demonstrate, this principled design, which effectively distills the ``inference-time finetuning" mechanism into a permanent set of parameters, not only leads to superior performance but also achieves significantly higher parameter efficiency.

% --- FIGURE for LoRA Mechanism ---
\section{Experiments}
\label{sec:experiments}

We conduct a comprehensive set of experiments to evaluate our proposed HiFICL paradigm. We first compare HiFICL against strong baselines to demonstrate its superiority (\cref{sec:main_results}). We then present in-depth ablation studies and analyses to validate our key design principles (\cref{sec:ablations}).

\subsection{Setup}
\label{sec:setup}

\paragraph{Models, Datasets, and Metrics.}
Our experiments are grounded on two popular open-source LMMs, LLaVA-Interleave-7b~\cite{liu2024LLaVAnext} and Idefics2-8b-base~\cite{laurenccon2024matters}. We specifically select these models as they both feature the fully autoregressive architecture upon which our theoretical analysis in \cref{sec:deconstruction} is based. Our evaluation spans three core multimodal benchmarks: VQAv2~\cite{goyal2017making}, OK-VQA~\cite{marino2019ok}, and COCO Captioning~\cite{chen2015microsoft}. Following the evaluation protocol of previous works~\cite{jiang2025mimic}, we randomly select 1000 samples per task for training. For evaluation, we report results on 10,000 validation samples from VQAv2 and the full validation splits for OK-VQA and COCO. We report VQA accuracy for the VQA tasks and CIDEr-D~\cite{vedantam2015cider} for captioning. All results are reported from the best-performing checkpoint.

\paragraph{Implementation Details.}
To ensure a fair comparison, our implementation details for all trainable baselines strictly follow those of MimIC~\cite{jiang2025mimic}. We employ the AdamW optimizer~\cite{loshchilov2017decoupled} with a learning rate of 5e-3, coupled with a cosine annealing scheduler with a warmup phase over the first 10\% of training steps. For HiFICL, we set the sequence length of the virtual matrices to $n=8$ (mimicking an 8-shot setting) across all experiments. We generally adopt a rank of $r=8$ as a robust default (with occasional adjustments to 4 or 16 depending on dataset complexity), leaving exact configurations detailed in the Appendix.

\subsection{Comparison with Existing Methods}
\label{sec:main_results}

\begin{table}[t]
\centering
\caption{Performance comparison on VQAv2, OK-VQA, and COCO. Best results are \textbf{bolded}, second best are \underline{underlined}.}
\label{tab:performance_comparison}
\footnotesize 
\setlength{\tabcolsep}{4.5pt} 
\renewcommand{\arraystretch}{1.2} 
\begin{tabular}{@{} >{\centering}m{1.2cm} l c c c c @{}}
\toprule
Model & Method & \# Params (M) & VQAv2 & OK-VQA & COCO \\
\midrule
\multirow{6}{*}{\rotatebox{90}{\textbf{LLaVA}}} 
    & Zero-shot     & - & 13.02      & 5.10       & 1.1516     \\
    & 8-shot ICL    & - & 68.19      & 43.84      & 1.2085     \\
    \cmidrule(lr){2-6} 
    & LoRA          & 19.7 ($\times$8.95) & 70.12 & 48.19 & 1.0665     \\
    & LIVE          & 0.13 ($\times$0.05) & 74.17      & 51.77      & 1.2743 \\
    & MimIC         & 17.0 ($\times$7.7) & \underline{74.40}      & \underline{52.29}      & \underline{1.3169}     \\
    & HiFICL        & 2.2 ($\times$1.0) & \textbf{74.66} & \textbf{54.19} & \textbf{1.3315} \\
\midrule
\multirow{8}{*}{\rotatebox{90}{\textbf{Idefics2}}}        
    & Zero-shot     & - & 55.39      & 43.08      & 0.6828     \\
    & 8-shot ICL    & - & 66.20      & 57.68      & 1.2119     \\
    & 16-shot ICL   & - & 67.70      & 57.96      & 1.2251     \\
    & 32-shot ICL   & - & 50.55      & 53.30      & 1.1806     \\
    \cmidrule(lr){2-6} 
    & LoRA          & 17.6 ($\times$8.0) & 69.14      & 55.05      & 1.2665     \\
    & LIVE          & 0.13 ($\times$0.06) & 67.60      & 54.86      & 1.2764 \\
    & MimIC         & 0.26 ($\times$0.12) & \underline{69.29} & \underline{58.74} & \underline{1.2827}    \\
    & HiFICL        & 2.2 ($\times$1.0) & \textbf{72.08} & \textbf{59.56} & \textbf{1.2951} \\
\bottomrule
\end{tabular}
\end{table}

\paragraph{Compared methods.}
We compare HiFICL against several categories of strong baselines. 
(1) \textbf{In-Context Learning (ICL)}: We evaluate the performance of Zero-shot and 8-shot ICL with randomly selected demonstrations, and additionally include 16-shot and 32-shot settings on the Idefics2 model for a broader ICL comparison.
(2) \textbf{PEFT}: We include LoRA~\cite{hu2022lora}, applying it with a rank of 16 to the attention matrices ($W_q, W_k, W_v, W_o$) to strictly follow the official setting of MimIC~\cite{jiang2025mimic} for a direct comparison.
(3) \textbf{ICL Approximation}: We benchmark against state-of-the-art methods from this paradigm which modify the language model part of the LMM. These methods differ in their insertion points: LIVE~\cite{peng2024live} introduces learnable vectors after each FFN layer, whereas both the prior state-of-the-art, MimIC~\cite{jiang2025mimic}, and our HiFICL operate directly within the Multi-Head Attention (MHA) block of each layer. 
For all trainable methods, we ensure a fair comparison by using the same training data and core hyperparameter settings.

% 【排版技巧】：跨栏大图提前声明，确保它能顺利排到下一页的顶部
\begin{figure*}[t]
  \centering
  \includegraphics[width=\linewidth]{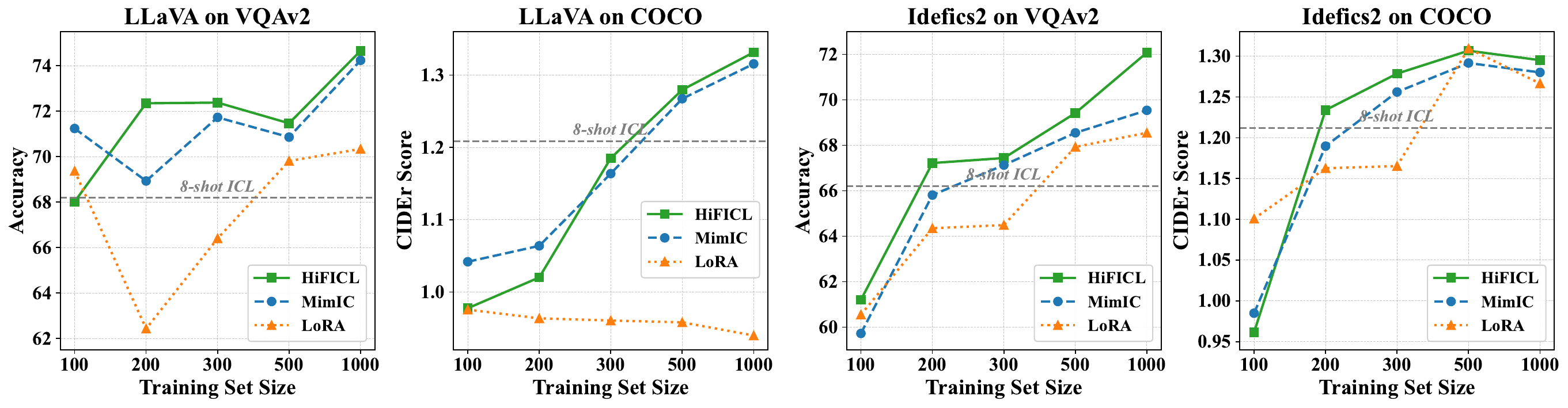}
  \caption{Data efficiency on VQAv2 and COCO across different models and training set sizes.}
  \label{fig:data_efficiency}
\end{figure*}

\paragraph{Results analysis.}
\cref{tab:performance_comparison} presents the results of HiFICL compared to all baselines. For the ICL baselines, 8-shot ICL significantly outperforms Zero-shot. On Idefics2, while 16-shot yields marginal gains over 8-shot at the cost of doubled computation, extending to 32-shot counterintuitively leads to performance degradation due to the excessively long context. For the trainable methods, nearly all approaches consistently improve over 8-shot ICL, with HiFICL demonstrating a commanding advantage. On the LLaVA-Interleave-7B model, HiFICL surpasses all competing methods, outperforming the prior best method, MimIC, by a substantial 1.9\% on OK-VQA and 1.1\% on COCO Captioning. Notably, the anomalous performance drop of LoRA on LLaVA for COCO is attributed to overfitting caused by the $r=16$ setting; a fair comparison with an optimized $r=8$ is detailed in the Supplementary Material. This superiority is further amplified on Idefics2, where the gap over MimIC widens to a remarkable 2.79\% on VQAv2. It is particularly noteworthy that HiFICL achieves this with a consistently modest parameter budget ($\sim$2.2M). Because MimIC trains a dense linear network $f(q)$, its parameter count scales heavily with the model's hidden dimension, exploding on high-dimensional models like LLaVA (17.0M). In contrast, our rank decomposition ensures HiFICL remains highly efficient, requiring roughly 8$\times$ fewer parameters than LoRA and MimIC on LLaVA while delivering superior results.

\paragraph{Data efficiency.}\label{par:data_efficiency}
We further analyze learning efficiency across five data scales (100 to 1000 samples). As illustrated in \cref{fig:data_efficiency}, HiFICL consistently demonstrates strong data efficiency. While minor performance fluctuations appear in the extreme low-data regime due to typical few-shot instability, HiFICL generally exhibits a steeper learning curve than competing methods. Notably, we observed that the LoRA baseline suffers from performance degradation in certain scenarios; this is primarily attributed to overfitting caused by the high rank setting ($r=16$) inherited from prior works. Despite this, HiFICL surpasses the strong 8-shot ICL performance with only around 300 training samples (e.g., on Idefics2 for COCO), suggesting it provides a more direct and effective learning signal.

\subsection{Ablations and More Analyses}
\label{sec:ablations}

% 【排版技巧】：消融实验的表2和图2紧跟在小节标题之后，排队进入双栏顶部
\begin{table}[t]
\centering
\caption{Ablation study of HiFICL's core components on the Idefics2 model.}
\label{tab:ablations}
\small 
\setlength{\tabcolsep}{5pt} 
\sisetup{detect-weight, mode=text}
\renewrobustcmd{\bfseries}{\fontseries{b}\selectfont}
\renewrobustcmd{\boldmath}{}
\begin{tabular}{
  @{}
  l 
  S[table-format=2.2]
  S[table-format=2.2]
  S[table-format=1.4]
  @{}
}
\toprule
\textbf{Method / Variant} & {\textbf{VQAv2}} & {\textbf{OK-VQA}} & {\textbf{COCO}} \\
\midrule
\textbf{HiFICL (Ours)} & \bfseries 72.08 & \bfseries 59.56 & \bfseries 1.2951 \\
\midrule
\multicolumn{4}{@{}l}{\textit{Training Paradigm:}} \\
\quad + Teacher & 70.09 & 59.13 & 1.2844 \\
\midrule
\multicolumn{4}{@{}l}{\textit{Dual LoRA Components:}} \\
\quad - LoRA on K & 70.58 & 55.72 & 1.2652 \\
\quad - LoRA on V & 69.31 & 56.86 & 1.2618 \\
\midrule
\multicolumn{4}{@{}l}{\textit{Non-linear Modulation:}} \\
\quad w/o SA scaling ($\alpha=1$) & 70.14 & 58.51 & 1.2808 \\
\bottomrule
\end{tabular}
\end{table}

\begin{figure}[t]
  \centering
  \includegraphics[width=\linewidth]{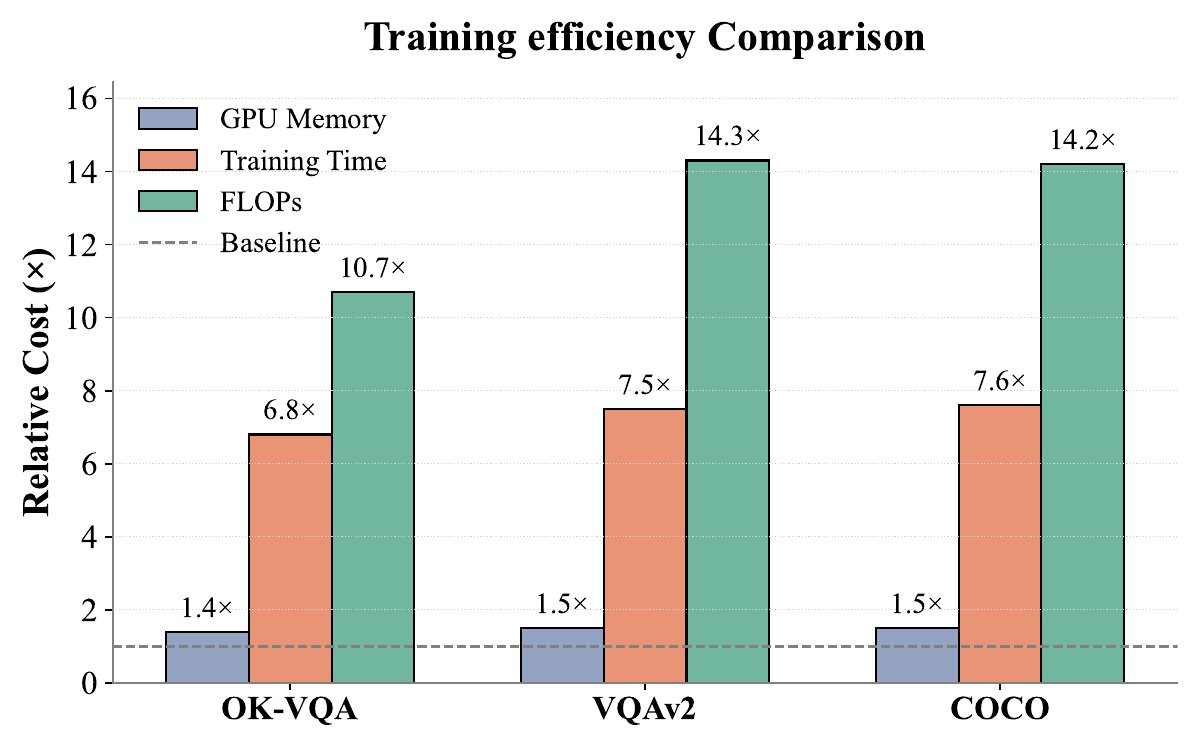} 
  \caption{Training efficiency comparison on the Idefics2 model. Bars show the relative cost of MimIC normalized against our method HiFICL (Baseline=1.0x). MimIC's paradigm incurs substantial overhead.}
  \label{fig:efficiency}
\end{figure}

We conduct a series of analytical experiments on the Idefics2-8b-base model to dissect the key components of HiFICL and validate our design choices. Our analysis confirms the necessity of each architectural component, demonstrates the substantial efficiency gains of our teacher-free paradigm, and explores the task-adaptive nature of our hyperparameters.

\paragraph{Core component ablations.}\label{par:core_ablations}
We systematically deconstruct HiFICL on the Idefics2 model to validate its key design principles, with results detailed in \cref{tab:ablations}. The findings confirm the necessity of each component. \textbf{First}, re-implementing HiFICL within a teacher-student framework (+ Teacher) by aligning hidden states similar to MimIC~\cite{jiang2025mimic} leads to a consistent performance drop (e.g., VQAv2 accuracy falls from 72.08\% to 70.09\%), supporting our hypothesis that the teacher model acts as a performance ceiling. \textbf{Second}, the dual low-rank decomposition is vital. While removing the factorization from either the key matrix (- LoRA on K) or the value matrix (- LoRA on V) significantly degrades performance, ablating $V$ incurs a more severe drop than ablating $K$ (e.g., 69.31\% vs. 70.58\% on VQAv2). This aligns perfectly with our theoretical derivation in \cref{sec:deconstruction}: $K$ primarily computes the dynamic mixture coefficients, whereas $V$ directly constitutes the contextual shift basis ($V_D$), making its parameterization more critical. \textbf{Finally}, disabling the self-attention scaling factor ($\alpha=1$), which reverts our method to a simpler linear shift approximation, also causes a substantial performance decline across all benchmarks. This empirically validates that capturing the full, non-linear attention dynamic is a key driver of HiFICL's superiority.

% 【排版技巧】：推理速度图提前在这里排队
\begin{figure}[t]
  \centering
  \includegraphics[width=\linewidth]{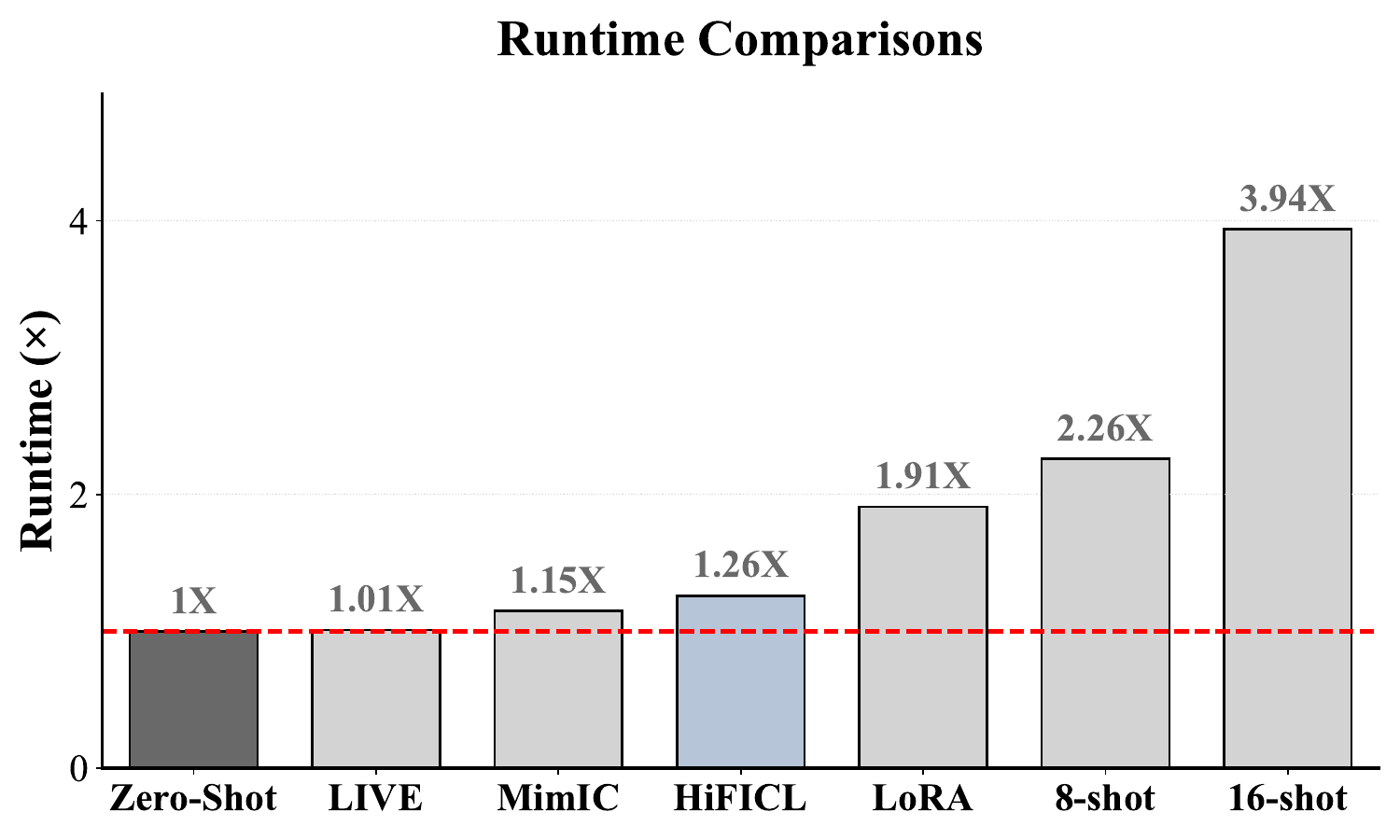} 
  \caption{Relative inference runtime comparison on VQAv2 using the LLaVA-Interleave-7b model. All runtimes are normalized to the Zero-shot baseline (1x).}
  \label{fig:inference_speed}
\end{figure}

\paragraph{Training efficiency.}\label{par:training_efficiency}
We compare the training cost of HiFICL against MimIC, which utilizes a teacher-student paradigm. As shown in \cref{fig:efficiency}, MimIC's approach incurs significant overhead, particularly in computation, with costs normalized against our teacher-free HiFICL (Baseline=1.0x). This large discrepancy arises because MimIC requires an additional, expensive forward pass through a large teacher model at every training step. Consequently, on VQAv2 for example, MimIC requires 7.5x the training time and a staggering 14.3x the FLOPs of HiFICL, while also using about 1.5x the GPU memory. Our end-to-end strategy completely obviates the costly forward pass, making HiFICL both more effective and significantly more practical.

% 【排版技巧】：Rank分析图提前在这里排队
\begin{figure}[t]
  \centering
  \includegraphics[width=\linewidth]{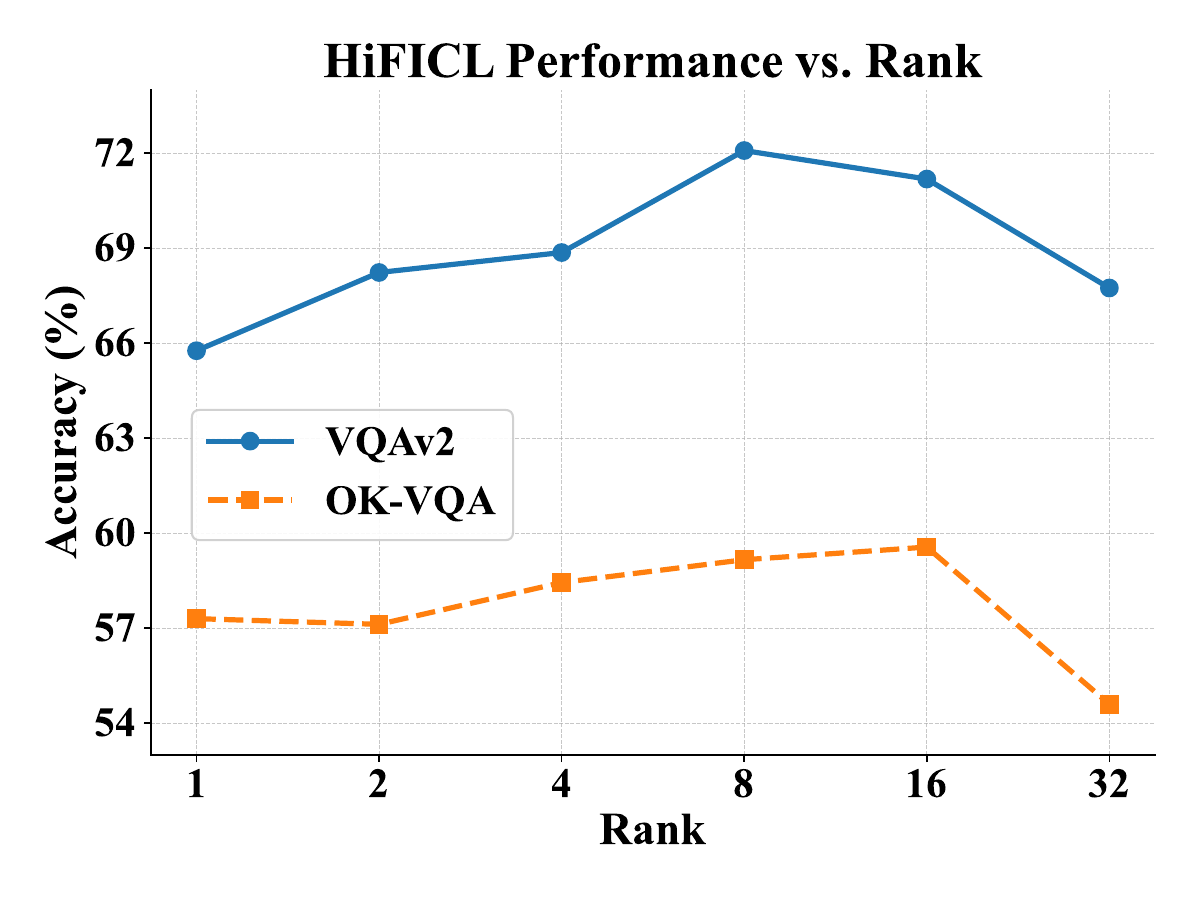} 
  \caption{Effect of rank ($r$) on HiFICL's performance across different datasets using the Idefics2 model. The optimal rank varies with task complexity, peaking at $r=8$ for VQAv2 and $r=16$ for OK-VQA.}
  \label{fig:rank_analysis}
\end{figure}

\paragraph{Inference efficiency.}\label{par:inference_efficiency}
A primary motivation for ICL approximation is to reduce the high inference cost of long contexts. We evaluate this by comparing the average processing time per sample on VQAv2 using the LLaVA-Interleave-7b model. As shown in \cref{fig:inference_speed}, all ICL approximation methods, including HiFICL, offer a dramatic speedup over standard multi-shot ICL. HiFICL's inference runtime is nearly on par with the Zero-shot baseline and comparable to prior approximation methods. Crucially, it is approximately 2.0x faster than 8-shot ICL and 3.4x faster than 16-shot ICL, confirming that our high-fidelity approach retains the core efficiency benefits of ICL approximation.

% 【排版技巧】：幻觉分析表提前在这里排队
\begin{table}[t] 
  \centering
  \caption{Hallucination analysis on COCO Captioning (Idefics2). Lower CHAIR is better, higher Recall is better.}
  \label{tab:hallucination}
  \footnotesize
  \setlength{\tabcolsep}{3pt}
  \begin{tabular}{@{}lcccccc@{}}
    \toprule
    & \textbf{0-shot} & \textbf{8-shot ICL} & \textbf{LoRA} & \textbf{LIVE} & \textbf{MimIC} & \textbf{HiFICL} \\
    \midrule
    \textbf{CHAIRs} $\downarrow$ & \textbf{2.8} & 5.6 & 3.3 & 4.4 & 4.0 & 3.2 \\
    \textbf{CHAIRi} $\downarrow$ & 3.3 & 3.9 & 2.4 & 3.1 & 2.9 & \textbf{2.2} \\
    \textbf{Recall} $\uparrow$   & 27.2 & 44.2 & 44.3 & 44.8 & 45.4 & \textbf{45.7} \\
    \bottomrule
  \end{tabular}
\end{table}

\paragraph{Task-adaptive rank.}\label{par:hyperparam_analysis}
Our analysis of the low-rank decomposition on Idefics2 reveals that the optimal rank $r$ correlates closely with task complexity. As shown in \cref{fig:rank_analysis}, the knowledge-intensive OK-VQA benefits from a higher capacity ($r=16$), whereas VQAv2 peaks earlier at $r=8$. Crucially, HiFICL shares rank-setting principles with LoRA. Empirically, $r=8$ emerges as a robust ``sweet spot'' across most scenarios, offering a reliable heuristic without extensive search. This demonstrates that our low-rank design acts not merely as a compression technique, but as a crucial, task-adaptive regularizer.

\paragraph{Hallucination analysis on image captioning.}\label{par:hallucination_analysis}
Beyond standard metrics, we evaluate content reliability by analyzing object hallucinations on COCO Captioning using the CHAIR metric~\cite{rohrbach2018object}. As reported in \cref{tab:hallucination}, HiFICL demonstrates a superior ability to produce faithful descriptions. It achieves the best (lowest) instance-level hallucination score (CHAIRi of 2.2), significantly outperforming all baselines, including 8-shot ICL (3.9), while simultaneously attaining the highest Recall (45.7). This suggests that our high-fidelity approach effectively reduces hallucinations without sacrificing descriptive detail, making it less prone to generating content ungrounded in the visual input.
\section{Conclusion}
\label{sec:conclusion}

In this paper, we introduced \textbf{Hi}gh-\textbf{F}idelity \textbf{I}n-\textbf{C}ontext \textbf{L}earning (\textbf{HiFICL}), a new paradigm for In-Context Learning (ICL) approximation. Our approach is grounded in a rigorous decomposition of the attention mechanism, which not only reframes the approximation problem by shifting the objective from modeling indirect \textit{effects} to directly parameterizing the \textit{source}, but also theoretically clarifies the underlying mechanism of ICL itself. We materialized this principle through two key innovations: a dual low-rank virtualization for high-fidelity approximation and a simple end-to-end training objective. Extensive experiments demonstrate that HiFICL consistently establishes a new state-of-the-art across multiple multimodal benchmarks, validating the superiority of our high-fidelity design.

Beyond ICL approximation, our work bridges the gap with Parameter-Efficient Fine-Tuning (PEFT). By practically instantiating the concept of ``inference-time finetuning,'' HiFICL demonstrates the viability of dynamic, content-aware adaptation. We hope this context-aware parameterization approach can serve as a useful reference for future developments in efficient model tuning.

% 使用 \clearpage 强制另起一页，并将之前未排版完的图表全部输出
\clearpage

\section*{Acknowledgment}
This work was supported by the National Natural Science Foundation of China under Grant No. 62472072.

{
    \small
    \bibliographystyle{ieeenat_fullname}
    \bibliography{main}
}

\clearpage
\clearpage
\setcounter{page}{1}
\maketitlesupplementary

\section{Implementation Details}
\label{sec:impl_details}

\subsection{Prompts}
\label{subsec:prompts}
We use unified prompt templates for VQAv2, OK-VQA, and COCO
to isolate the effect of the adaptation method from prompt
engineering. For VQA tasks we adopt an instruction-style
prefix that asks the model to answer a question given an
image; for COCO we use a caption-style prefix.

For all ICL settings (0-shot, 8-shot, LoRA, LIVE, MimIC, and HiFiICL),
the same templates are used for every in-context
demonstration; only the number of demonstrations and their
ordering differ. \cref{tab:supp_prompts} lists the exact
prefixes, ICL patterns, and decoding stop words for each
task, matching the configurations used in the main paper.

\begin{table*}[t]
  \centering
  \caption{Prompt templates used in our experiments. Curly brackets \{\} indicate fields filled with instance-specific content.}
  \label{tab:supp_prompts}
  \begin{tabular}{@{} l p{0.27\textwidth} p{0.34\textwidth} p{0.16\textwidth} @{}}
    \toprule
    Task & Prefix prompt & ICL prompt & Stop words \\
    \midrule
    VQAv2 / OK-VQA
      & Instruction: provide an answer to the question. Use the image to answer.
      & Image: \{image\} Question: \{question\} Answer: \{answer\}\textbackslash n
      & ``Question'', ``Answer'', ``Image'' \\
    \midrule
    COCO
      & ---
      & \multirow{2}{\linewidth}{Image: \{image\} Caption: \{caption\}\textbackslash n}
      & \multirow{2}{\linewidth}{``Caption'', ``Image''} \\
    COCO ICL
      & Instruction: provide a short caption of the input image.\textbackslash n
      &  &  \\
    \bottomrule
  \end{tabular}
\end{table*}

\subsection{Hyperparameters}
\label{subsec:hyperparams}

Our hyperparameter settings for all trainable methods are designed for a fair and rigorous comparison, strictly following the configurations from their respective original papers where applicable. A summary of the main hyperparameters is provided in \cref{tab:supp_hparams}.

For our method, \textbf{HiFICL}, and for \textbf{MimIC}~\cite{jiang2025mimic}, we adopt the same core training configuration. We use an AdamW optimizer~\cite{loshchilov2017decoupled} with a learning rate of 5e-3, coupled with a cosine annealing scheduler with a 10\% warmup phase.

For \textbf{LIVE}~\cite{peng2024live}, the base learning rate is also 5e-3. However, in accordance with the original paper, the separate learning rate for its shift magnitude parameter is set to 1e-2.

For \textbf{LoRA}~\cite{hu2022lora}, we set the rank to $r=16$. Given its substantial parameter count, we use a lower learning rate of 5e-4 to ensure stable training.

For all methods, when the training set size is 1000, we perform training for 5 epochs. For the smaller datasets in our data efficiency analysis, this is increased to 10 epochs. The task-specific ranks for HiFICL are detailed in \cref{tab:supp_hifi_rank}.

\begin{table}[h]
  \centering
  \caption{Main hyperparameters for all trainable methods. Specific learning rates for LoRA and LIVE's magnitude parameter are noted in the text.}
  \label{tab:supp_hparams}
  \begin{tabular}{lc}
    \toprule
    Hyperparameter & Value \\ 
    \midrule
    Optimizer & AdamW \\
    Base Learning Rate & 5e-3 \\
    LR Schedule & Cosine w/ 10\% Warmup \\
    Weight Decay & 0.05 \\
    Precision & 16-mixed \\
    Batch Size (per GPU) & 2 \\
    Gradient Accumulation & 2 \\
    Total Training Epochs & 5 (for 1k) / 10 ($<$1k) \\
    \bottomrule
  \end{tabular}
\end{table}

\begin{table}[h]
  \centering
  \caption{The optimal rank $r$ used in HiFICL for each dataset and backbone model.}
  \label{tab:supp_hifi_rank}
  \begin{tabular}{lcc}
    \toprule
    Dataset & LLaVA-Interleave & Idefics2 \\
    \midrule
    VQAv2    & 8 & 8 \\
    OK-VQA   & 4 & 16 \\
    COCO     & 8 & 4 \\
    \bottomrule
  \end{tabular}
\end{table}

\section{Backbone Models and Qualitative Analysis}
\label{sec:backbone_qual}

\subsection{Why Idefics2 and LLaVA-Interleave?}
\label{subsec:why_two_backbones}

We selected two open-source LMMs for our experiments:

\begin{itemize}
    \item \textbf{Idefics2-8b-base:} An official open-source model from Hugging Face, serving as a standard and reproducible testbed for multimodal research. Its architecture is a clean, fully autoregressive design.
    \item \textbf{LLaVA-Interleave-7b:} A model from the LLaVA-NeXT series, representing one of the most mainstream and widely-used families of open-source LMMs. It is also one of the few LLaVA models that natively supports the interleaved image-text inputs required for ICL studies.
\end{itemize}

The choice of these two models allows for a robust evaluation of HiFICL's generalization. Idefics2 is a pre-trained backbone, ideal for analyzing fundamental ICL behavior. LLaVA-Interleave is an instruction-tuned model, representing a different and highly optimized training paradigm. Our results also show they have different performance profiles: Idefics2 performs better on VQAv2 and COCO, while LLaVA-Interleave excels on OK-VQA. Demonstrating strong performance on both validates HiFICL's broad applicability.

Practically, both 7-8B scale models fit on a single 80GB A100 GPU for all experimental conditions, including ICL, LoRA, and all approximation methods. This ensures a controlled and fair comparison environment by eliminating the complexities of multi-GPU setups.

\subsection{Why Not Other Backbone Models?}
\label{subsec:why_not_other_backbones}

To further justify our model selection, we also evaluated HiFICL on three alternative backbones. The findings, summarized in \cref{tab:backbone_comparison}, reinforce our choice of Idefics2 and LLaVA-Interleave for the main experiments.

On the \textbf{Idefics1-9B}, which utilizes a cross-attention architecture, HiFICL does not show a clear advantage over MimIC. This is consistent with our theoretical framework, as our method's design is grounded in the mathematical properties of fully autoregressive self-attention mechanisms.

The case of \textbf{LLaVA-v1.6-Mistral-7B} is particularly insightful. This model is not natively designed for ICL, and standard 8-shot ICL degrades its performance. However, both HiFICL and MimIC successfully adapt the model and significantly improve over the zero-shot baseline, demonstrating the robustness of ICL approximation methods in this scenario.

Finally, while \textbf{LLaVA-OneVision-Qwen2-7B} shows better performance than LLaVA-Interleave, its design for long video sequences results in a doubled GPU memory requirement due to a larger hidden state size. Given that both models are based on the same Qwen2 language model and their performance on our image-based tasks is largely comparable, we selected LLaVA-Interleave for our main experiments. This choice prioritizes a more favorable balance between performance and computational cost, ensuring better reproducibility for the research community.

\begin{table}[h]
  \centering
  \small
  \setlength{\tabcolsep}{4pt}
  \caption{Performance comparison across different backbone models. Best results in each group are \textbf{bolded}, second best are \underline{underlined}.}
  \label{tab:backbone_comparison}
  \begin{tabular}{@{}llccc@{}}
    \toprule
    Backbone & Method & VQAv2 & OK-VQA & COCO \\
    \midrule
    \multirow{4}{*}{Idefics1-9B} 
        & Zero-shot & 29.25 & 30.54 & 63.06 \\
        & 32-shot ICL& 56.18 & 48.48 & 105.89 \\
        & MimIC     & \underline{59.64} & \textbf{52.05} & \underline{114.89} \\
        & HiFICL    & \textbf{59.71} & \underline{51.93} & \textbf{115.21} \\
    \midrule
    \multirow{4}{*}{\shortstack{LLaVA-v1.6 \\ (Mistral-7B)}}
        & Zero-shot & 70.00 & 63.00 & 0.7157 \\
        & 8-shot ICL& 68.00 & 56.00 & 0.6678 \\
        & MimIC     & \underline{71.24} & \underline{64.62} & \underline{1.2857} \\
        & HiFICL    & \textbf{74.71} & \textbf{66.88} & \textbf{1.3192} \\
    \midrule
    \multirow{4}{*}{\shortstack{LLaVA-OneVision \\ (Qwen2-7B)}}
        & Zero-shot & 71.75 & 48.19 & 1.2091 \\
        & 8-shot ICL& 78.70 & 66.59 & 1.3457 \\
        & MimIC     & \underline{81.22} & \underline{69.43} & \underline{1.4312} \\
        & HiFICL    & \textbf{82.39} & \textbf{73.12} & \textbf{1.4784} \\
    \bottomrule
  \end{tabular}
\end{table}

\subsection{Additional Qualitative Examples}
\label{subsec:qualitative_more}

\begin{figure*}[t]
  \centering
  \includegraphics[width=\textwidth]{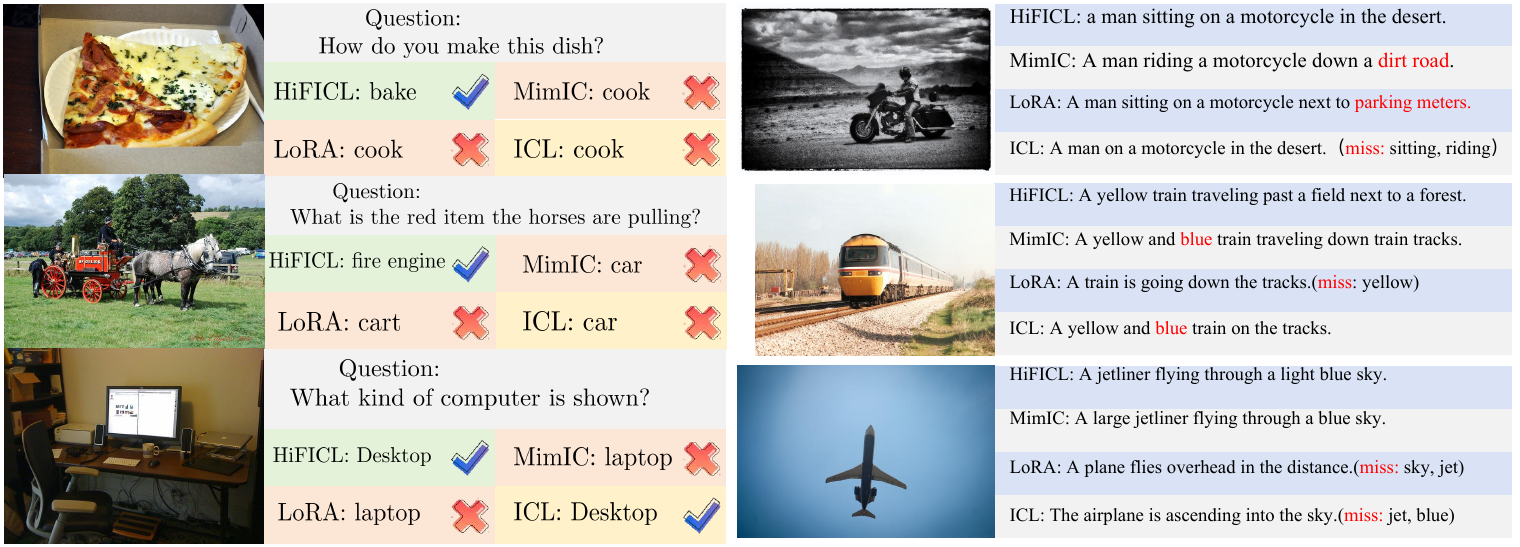} % <<< CHANGE THIS PATH
  \caption{Qualitative examples on VQAv2 (left) and COCO Captioning (right) using the LLaVA-Interleave-7b model.}
  \label{fig:qual_vqa_supp}
\end{figure*}

We provide qualitative examples on VQAv2 and COCO in \cref{fig:qual_vqa_supp} to complement our quantitative results. The examples, generated using the LLaVA-Interleave-7b model, visually demonstrate HiFICL's ability to produce more faithful responses. For instance, HiFICL often corrects factual errors made by other methods (e.g., identifying a ``fire engine'' instead of a ``car'') and reduces object hallucination (e.g., avoiding the erroneous ``parking meters''), showcasing its effectiveness in capturing nuanced visual details.

\section{LoRA Variants and Efficiency Analysis}
\label{sec:lora_efficiency}

\subsection{Overview of Recent LoRA Variants}
\label{subsec:lora_overview}

\begin{figure}[t]
  \centering
  \includegraphics[width=\linewidth]{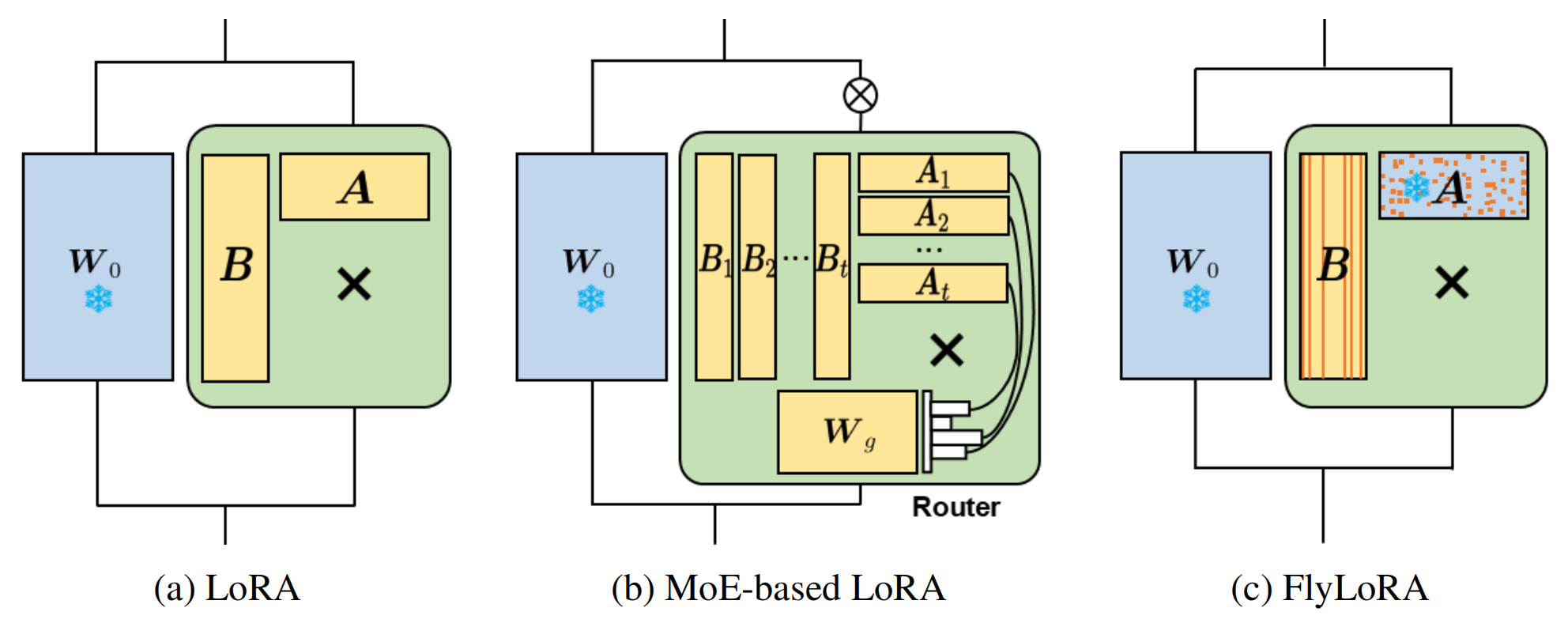} % <<< 修改了 width
  \caption{Schematic illustrations of different LoRA variants. 
(a) Standard LoRA applies a dense low-rank update using matrices $A$ and $B$. 
(b) MoE-based LoRA decomposes the update into multiple smaller "expert" pairs $\{A_i, B_i\}$ and uses a router to select a sparse combination. 
(c) FlyLoRA utilizes a frozen random matrix for $A$ and only trains a sparse matrix $B$, activating a small subset of its columns for each input.}
  \label{fig:lora_family}
\end{figure}

LoRA-style PEFT has evolved rapidly. As shown in Figure~\ref{fig:lora_family}, variants like MoE-based LoRA or FlyLoRA innovate the low-rank update matrix ($\Delta W$). However, they all operate in the static \textbf{weight space} via additive updates. 

HiFICL shifts this paradigm: instead of modifying the weight space, it operates in the dynamic \textbf{activation space} to directly model the ICL-induced shift. Because HiFICL also employs low-rank virtual matrices, existing LoRA structural innovations are potentially complementary. For instance, a future ``MoE-HiFICL'' could dynamically route to multiple virtual key-value pairs, opening new avenues for context-aware PEFT.

% --- 修改点 1：把图片稍微缩小到 0.9\linewidth 腾出垂直空间 ---

\subsection{Efficiency Comparison: HiFICL vs. LoRA}
\label{subsec:hificl_vs_lora_efficiency}

\begin{figure}[t]
  \centering
  \includegraphics[width=0.9\linewidth]{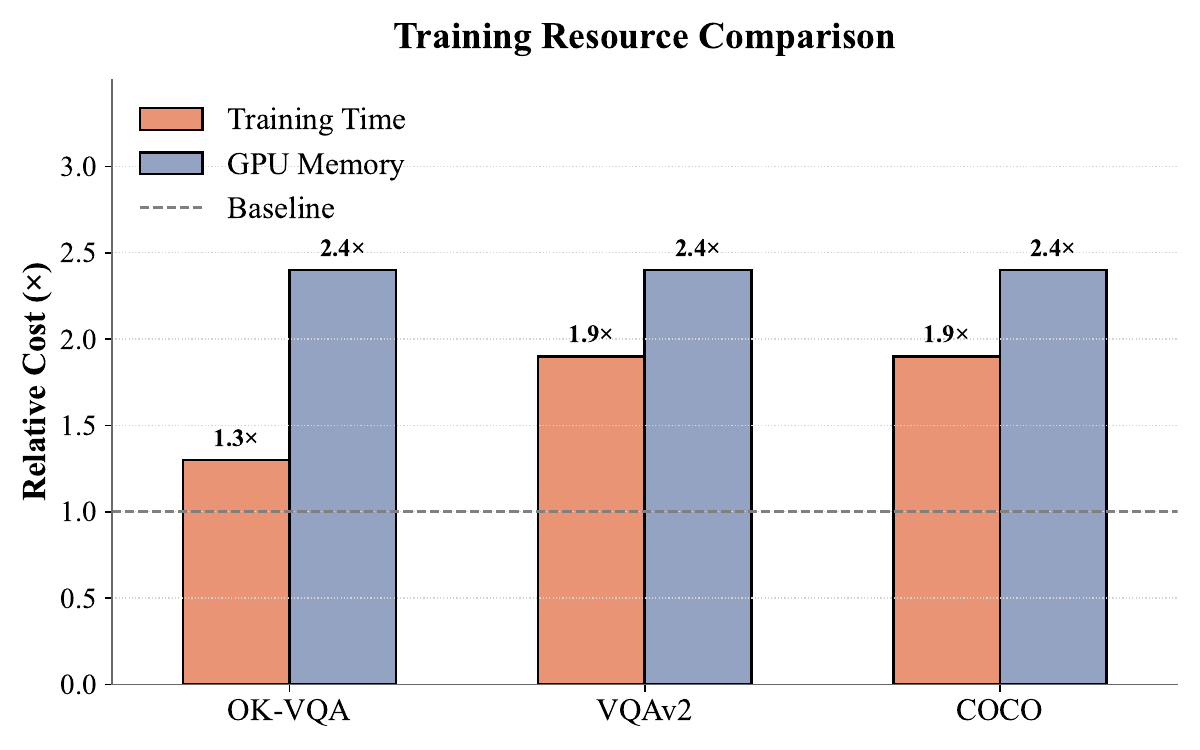} % <<< 修改了 width
  \caption{Training efficiency comparison between LoRA and HiFICL on the Idefics2 model. Costs (GPU Memory, Training Time, FLOPs) are relative to HiFICL (Baseline=1.0x).}
  \label{fig:hificl_vs_lora_training}
\end{figure}

While both HiFICL and LoRA are parameter-efficient, their distinct mechanisms lead to fundamental differences in training efficiency. LoRA's updates are applied to the core weight matrices ($W_q, W_k$, etc.). To compute gradients for its adapter matrices ($A, B$), it requires a full backpropagation pass through the entire frozen backbone, making the process computationally intensive. In contrast, HiFICL injects its parameters into the activation space. This architectural choice enables a more localized and efficient gradient computation, as the backpropagation path for the trainable virtual key-value pairs does not involve the main backbone. As empirically verified in Figure~\ref{fig:hificl_vs_lora_training}, this principled design consistently results in lower training time and GPU memory consumption compared to LoRA across all evaluated tasks, establishing HiFICL as a more computationally efficient PEFT solution for training.

% --- 修改点 2：把图片稍微缩小到 0.9\linewidth 腾出垂直空间 ---

\subsection{Performance at Optimal Rank ($r=8$)}
\label{subsec:optimal_rank_comparison}

As discussed in the main text, the default rank setting ($r=16$) inherited from prior ICL approximation baselines can lead to overfitting for LoRA on certain datasets. To provide a generic and fair PEFT comparison, we evaluate both LoRA and HiFICL using an optimized rank of $r=8$, which empirically serves as a robust ``sweet spot'' for both methods. 

As shown in \cref{tab:lora_optimal_rank}, when standardizing the rank to $r=8$, the performance degradation of LoRA vanishes, confirming that the earlier drop was indeed due to capacity-induced overfitting. In this optimized setting, HiFICL achieves performance parity with LoRA across both LLaVA and Idefics2 models. Crucially, HiFICL achieves this highly competitive performance while utilizing approximately 4$\times$ fewer trainable parameters (2.2M vs. 8.8M/9.8M). This demonstrates that our context-aware parameterization approach is not only mathematically principled but also highly parameter-efficient compared to standard weight-space adaptation.

% --- 修改点 3：把 [h] 改成 [htbp] ---
\begin{table}[htbp] % <<< 这里非常关键，去掉了原先死板的 [h]
\centering
\caption{Performance comparison between LoRA and HiFICL with an optimal rank of $r=8$. Best results for each model are \textbf{bolded}.}
\label{tab:lora_optimal_rank}
\footnotesize 
\setlength{\tabcolsep}{4.5pt} 
\renewcommand{\arraystretch}{1.2} 
\begin{tabular}{@{} >{\centering}m{1.2cm} l c c c c @{}}
\toprule
Model & Method & \# Params (M) & VQAv2 & OK-VQA & COCO \\
\midrule
\multirow{2}{*}{\textbf{LLaVA}} 
    & LoRA          & 9.8 ($\times$4.5) & \textbf{75.75} & \textbf{54.28} & 1.3307 \\
    & HiFICL        & 2.2 ($\times$1.0)  & 74.66          & 53.19          & \textbf{1.3315} \\
\midrule
\multirow{2}{*}{\textbf{Idefics2}}       
    & LoRA          & 8.8 ($\times$4.0)  & 69.58          & \textbf{60.18} & \textbf{1.3448} \\
    & HiFICL        & 2.2 ($\times$1.0)  & \textbf{72.08} & 58.96          & 1.2851 \\
\bottomrule
\end{tabular}
\end{table}

\end{document}